\documentclass[10pt,twocolumn,letterpaper]{article}

\usepackage[final]{cvpr}      

\usepackage[table,xcdraw,dvipsnames]{xcolor}

\definecolor{cvprblue}{rgb}{0.21,0.49,0.74}
\usepackage[pagebackref,breaklinks,colorlinks,citecolor=cvprblue]{hyperref}

\usepackage{makecell}

\usepackage[accsupp]{axessibility} 

\definecolor{mygray}{gray}{0.95}
\definecolor{myblue}{rgb}{0.863, 0.902, 0.945}
\definecolor{myred}{rgb}{0.949, 0.863, 0.859}

\graphicspath{{fig}}

\def\paperID{15} 
\def\confName{CVPR}
\def\confYear{2024}

\title{Towards Real-world Video Face Restoration: A New Benchmark}

\author{
Ziyan Chen$^{1,3}$\thanks{Equal Contribution. 
}, Jingwen He$^{2,3*}$, Xinqi Lin$^{1}$, Yu Qiao$^{3}$, Chao Dong$^{1,3}$\thanks{Corresponding author.} \\
{$^{1}$Shenzhen Institute of Advanced Technology, Chinese Academy of Sciences} \\
{$^{2}$The Chinese University of Hong Kong} \\
{$^{3}$Shanghai AI Laboratory}\\
{\tt\small 
\{chen.ziyan, hejingwenhejingwen, linxinqi007\}@outlook.com, 
\{dongchao, qiaoyu\}@pjlab.org.cn
}
}

\begin{document}
\maketitle
\begin{abstract}
Blind face restoration (BFR) on images has significantly progressed over the last several years, while real-world video face restoration (VFR) remains unsolved. One current challenge in the VFR task is the absence of an effective evaluation system. The real-world video face test dataset is lacking, and the evaluation metrics require validation and supplementation before applying to video scenarios. 
In this work, we introduced new real-world datasets named \textbf{FOS} with a taxonomy of ``\textit{\textbf{F}ull}, \textit{\textbf{O}ccluded}, and \textit{\textbf{S}ide}" faces derived from mainly video frames to study the applicability of current methods on videos. 
Compared with existing test datasets, FOS datasets cover real-world video test sets, more diverse degradations, and complex scenarios. 
Given the established datasets, we benchmarked both the state-of-the-art BFR methods and the video super resolution (VSR) methods to comprehensively study current approaches, identifying their potential and limitations in VFR tasks. 
In addition, we studied the effectiveness of the commonly used image quality assessment (IQA) metrics and face IQA (FIQA) metrics by leveraging a subjective user study.
The extensive experimental results and detailed analysis pose challenges to current face restoration approaches, which we hope stimulate future advances in VFR research. 
The project page can be found in \href{https://ziyannchen.github.io/projects/VFRxBenchmark/}{https://ziyannchen.github.io/projects/VFRxBenchmark/}.

\end{abstract}    
\section{Introduction}
\begin{figure} 
    \centering
    \subfloat[Restoration results on commonly used real-world image datasets.\label{subfig1:previous_samples}]{%
        \includegraphics[width=0.95\linewidth]{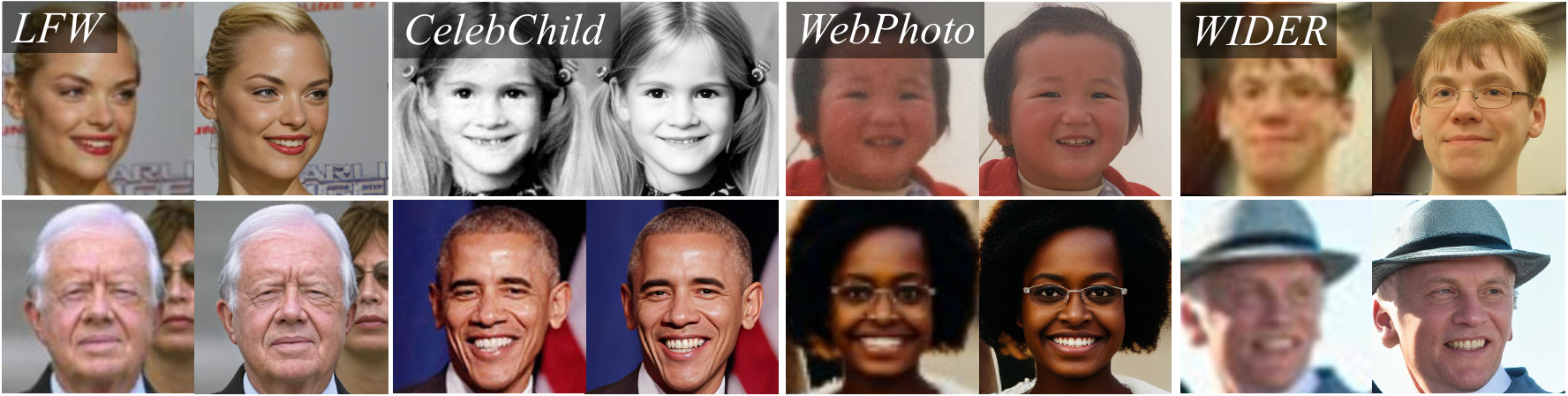}}
    \\
    \subfloat[Restoration failures on our proposed FOS datasets.\label{subfig1:fos_demo}]{%
        \includegraphics[width=0.95\linewidth]{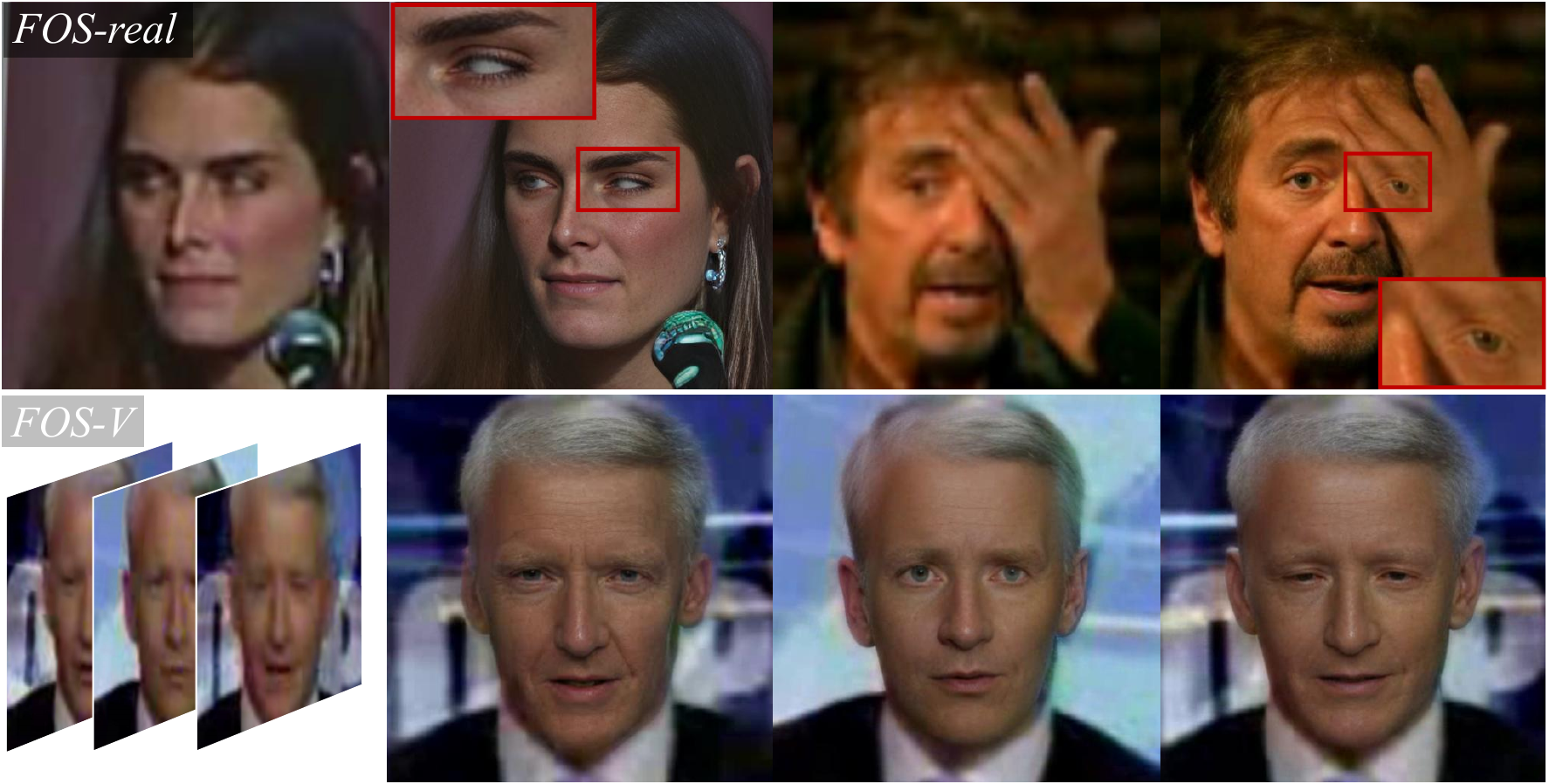}}
    \vspace{-0.5em}
    \caption{The face restoration results achieved by CodeFormer \cite{codeformer} on widely used real-world datasets (a) \cite{gfpgan, codeformer} and our proposed FOS datasets (b).  (\textbf{Zoom in for details})}
  \label{fig:corr_quality}
  \vspace{-1em}
\end{figure}
Blind face restoration (BFR) aims at recovering a realistic and faithful face image from a corrupted one where the specific degradation is unknown.
It is a challenging problem as the real-world scenarios contain complicated degradations and diverse facial poses as well as expressions. 
Existing blind face restoration works \cite{gfpgan, gpen, gcfsr, restoreformer, vqfr, codeformer} have achieved unprecedented progress by incorporating powerful generative models as facial priors, such as generative adversarial network (GAN) \cite{gan, stylegan1, stylegan2, stylegan3} and vector-quantized autoencoder \cite{vqvae, vqvae2, vqgan}. 
These blind face restoration algorithms \cite{gfpgan, gpen, codeformer} have shown a huge impact in the open source community and significant commercial potential in image enhancement applications.

\begin{figure*} 
    \centering
    \subfloat[\textbf{FOS-syn}. The FOS-syn dataset contains a total number of 3, 150 face images derived from widely-used CelebA-HQ~\cite{celebahq}.\label{fig:overview_fos_syn}]{%
        \includegraphics[width=1\linewidth]{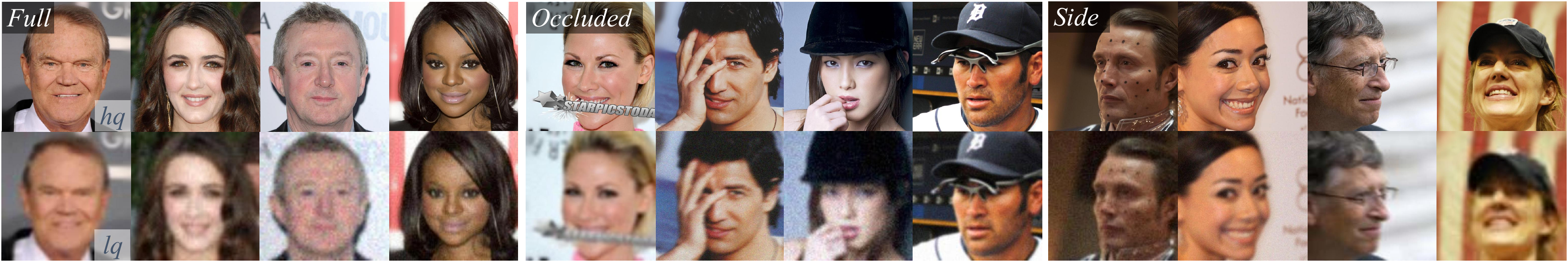}}
    \\ \vspace{0mm}
    \subfloat[\textbf{FOS-real}. The FOS-real dataset involves a total number of 4, 253 face images extracted from face videos.\label{fig:overview_fos_real}]{%
        \includegraphics[width=1\linewidth]{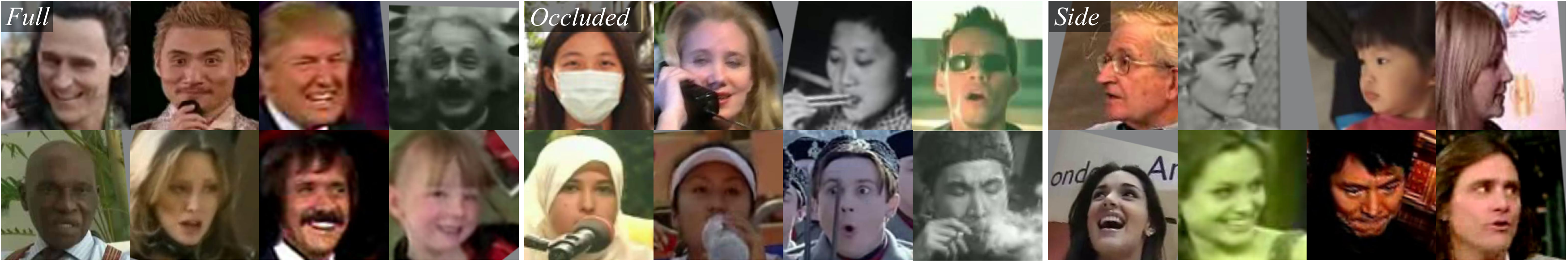}}
    \\ \vspace{0mm}
    \subfloat[\textbf{FOS-V}. The FOS-V dataset consists of 3,316 face clips originating from YouTube videos.\label{fig:overview_fos_v}]{%
        \includegraphics[width=1\linewidth]{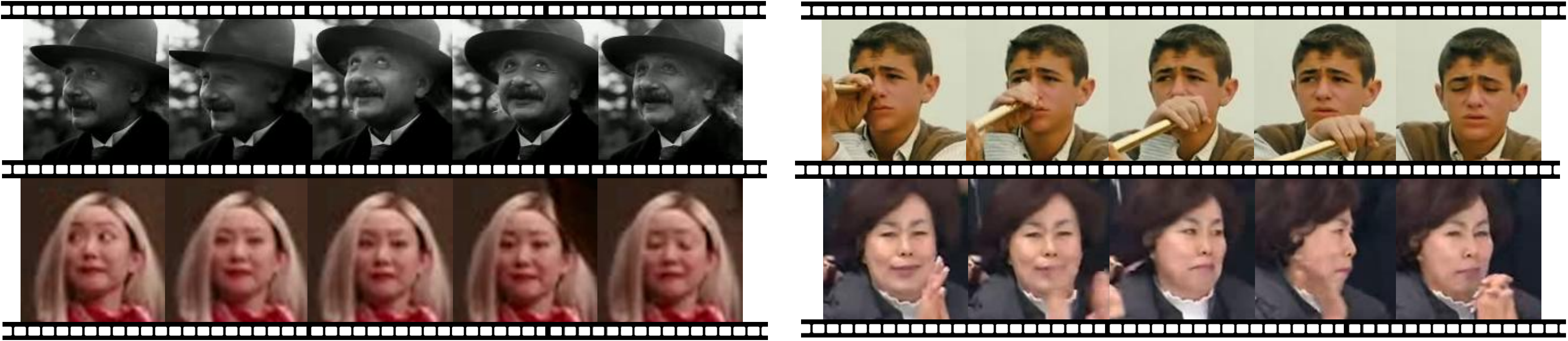}}
        \vspace{-2mm}
    \caption{Overview of the proposed FOS datasets.}
  \label{fig:overview_fos}
  \vspace{-1em}
\end{figure*}

Despite promising progress in BFR, the video face restoration (VFR) remains unsolved. Although a few methods have been proposed in face super-resolution~\cite{zou2023flair, vfsr1, vfsr2, vfsr3}, there is still a huge gap between the performances of BFR and VFR methods. 
Figure \ref{subfig1:previous_samples} shows that state-of-the-art BFR methods can yield incredible results in both identity preservation and facial features/texture details regeneration on widely used real image testing datasets (\eg, WebPhoto-Test~\cite{gfpgan}, LFW-Test~\cite{lfw_dataset}, Wider-Test~\cite{codeformer}, CelebChild~\cite{gfpgan}). However, these commonly used test datasets cover limited scenarios. Specifically, WebPhoto-Test and Wider-Test are mainly about human photography, where (1) most faces are assured to have no occlusion, (2) diversity of the gaze and head directions are lacking, and (3) the facial expressions are not diverse enough. Although these missing features may be marginal in the image cases, they are commonly found in video cases.
Figure \ref{subfig1:fos_demo} indicates obvious failures of the most recent state-of-the-art BFR method CodeFormer \cite{codeformer} in restoring side/occluded faces or maintaining stability in face video frames. 
 
These observations reveal that a complete evaluation system applicable to the VFR task is missing. Real video face test sets remain lacking, and limitations of the existing image test datasets are not fully reflected. Therefore, we introduce a new benchmark in this work, namely \textbf{FOS}, hoping to advance future VFR research. Specifically, we construct three test datasets, \textbf{FOS-V}, \textbf{FOS-real} and \textbf{FOS-syn}, with three representative face poses/conditions highlighted as \textit{\textbf{F}ull}, \textit{\textbf{O}ccluded}, and \textit{\textbf{S}ide}. 
An overview of our FOS datasets is presented in Table \ref{tab:overview}, while visualized samples are in Figure \ref{fig:overview_fos}.
Compared to the existing test sets, FOS datasets show larger diversity that involves heterogeneous scenes (\eg interviews, sports, concerts, and old movies) and a wide range of faces with different facial expressions, poses, and ethnicities.
 
Given the collected FOS testing datasets, we benchmark 11 representative state-of-the-art BFR methods and 4 VSR implementations \cite{vfhq} to reveal their potential and limitations for the FVR task.
In addition, we empirically discovered that the commonly used metrics are inconsistent with the human eye. We thus explore the effectiveness of current image quality assessment (IQA) metrics and the face IQA (FIQA) by conducting a subjective user study based on the benchmarking results.
This work aims to stimulate future advances in video face restoration. The contributions of this work can be summarized below. 
\begin{enumerate}
    \item We introduce a comprehensive real-world face test dataset, involving a video test subset to fill a gap in the VFR field where real test sets are missing.
    \item By benchmarking the state-of-the-art BFR and VSR methods, we identify future challenges in FVR field. None of the existing methods can achieve satisfying intra-frame reconstruction performance while maintaining inter-frame stability. 
    \item We leverage a user study to investigate the effectiveness of IQA and FIQA metrics. The results show that NIQE~\cite{niqe} and BRISQUE~\cite{brisque} are significantly inconsistent with subjective scores, while FIQA~\cite{serfiq, faceqnet, sddfiq} metrics yield a strong correlation with the human eye. 
    \item A new metric VIDD is introduced to assess the inter-frame stability of videos. 
\end{enumerate}


\section{Related Work}\label{sec:related-work}

\noindent\textbf{Blind face restoration}.
Recent works exploit multiple facial priors to help the blind restoration problem. These prior-based methods can be three mainstreams: 1) Geometric priors. Spatial information such as facial parsing maps~\cite{ChaofengChen2023ProgressiveSS}, facial landmarks~\cite{fsrnet, kim2019progressive}, and facial component heatmaps~\cite{yu2018face} are used to provide additional information on face shape and details. 2) Reference priors. Typically, a reference face or facial component dictionaries obtained from high-quality faces~\cite{XiaomingLi2018LearningWG, XiaomingLi2020EnhancedBF, XiaomingLi2020BlindFR, BerkDogan2019ExemplarGF} are leveraged to guide the face recovery. 3) Generative priors. A pretrained generative network, \eg StyleGAN~\cite{stylegan1, stylegan2} or VQ-GAN~\cite{vqgan}, is employed to provide realistic facial information~\cite{gfpgan, codeformer, restoreformer, vqfr}. 
GFP-GAN~\cite{gfpgan} and GPEN~\cite{gpen} made the first endeavor to extract fidelity information from low-quality input face images to balance the realness and fidelity. In addition, CodeFormer~\cite{codeformer}, RestoreFormer~\cite{restoreformer}, and VQFR~\cite{vqfr} exploit to fuse high-quality priors from the learned vector-quantized dictionary and information from the low-quality inputs to enable the discovery of natural and realistic faces that well approximate the target faces.

\noindent\textbf{Video restoration} 
Although BFR methods have achieved great success in single image restoration, video face restoration draws little attention from researchers. A few attempts have been made in video face super resolution~\cite{vfsr1, vfsr2, vfsr3, zou2023flair}, most of which focus on a fusion of inter-frame spatial and temporal information, or aural and visual modalities. Recently, \cite{zou2023flair} made a new attempt to apply the diffusion model to the VFR task. It is also worth noting that some modest progress has been made in general video super resolution. For example, EDVR~\cite{edvr} and BasicVSR~\cite{basicvsr} demonstrated their appealing performance in both temporal consistency and single-frame restoration by leveraging temporal alignment, and aggregation. 

\noindent\textbf{Datasets}
CelebA-Test~\cite{celebahq} is a widely used synthetic test set for single-image blind face restoration. The commonly-used testing datasets in the real-world scenarios, LFW-Test~\cite{lfw_dataset}, WIDER-Test~\cite{codeformer}, WebPhoto-Test~\cite{gfpgan} and CelebChild-Test~\cite{gfpgan} are either collected from the Internet or originated from other face-related tasks such as face verification or face detection. The face images of these testing datasets tend to be photos with a frontal view. However, the absence of more hard testing samples prevents evaluations from fully reflecting the models' performance in the real world. Furthermore, a standard and real-world video face testing set is still missing. VFHQ~\cite{vfhq} dataset is proposed as an alternative to VoxCeleb~\cite{voxceleb1, voxceleb2} with higher-quality face images and enables an improvement in VSR methods. 

\begin{table*}[htbp]
\vspace{-2em}
\centering
\footnotesize
\setlength\tabcolsep{3pt}
    \caption{Overview of the existing public real-world testing datasets and our FOS datasets.}
    \vspace{-1em}
    \begin{tabular}{ccccccc}
\toprule
\textbf{Dataset} & \textbf{\# Full} & \textbf{\# Occluded} & \textbf{\# Side} & \textbf{\# Total} & \textbf{ Real/synthetic} & \textbf{Descriptions}                                                               \\ \midrule
CelebChild  \cite{gfpgan}     & 325              & 27                    & 8                & 360               & real                 & childhood and recent photos of celebrities                                          \\
LFW-Test  \cite{gfpgan}       & 1566             & 113                   & 32               & 1711              & real                 & snapshots of celebrities                                                                     \\
WebPhoto-Test \cite{gfpgan}   & 383              & 19                    & 5                & 407               & real             & old photos                                                                 \\
WIDER-Test \cite{codeformer}      & 894              & 58                    & 18               & 970               & real                & group photos                                                                  \\
VFHQ-Test \cite{vfhq}       & -                & -                     & -                & 50                &  synthetic    & videos of interviews                                                                    \\ \hline \hline
FOS-syn  (Ours)        & 1020             & 1097                  & 1032             & 3150              & synthetic        & photos of celebrities                                                         \\ 
FOS-real  (Ours)       & 1633             & 1136                  & 1484             & 4253              & real        & images of sports, old movies, interviews, concerts, \etc. \\
FOS-V (Ours)           & -                & -                     & -                & 3316              & real        & videos of sports, old movies, interviews, concerts, \etc. \\

\bottomrule
\end{tabular}
    \vspace{-1em}
    \label{tab:overview}
\end{table*}

\section{Benchmark Settings}\label{sec:benchmark}
In this section, we introduce the established benchmark dataset and our evaluation settings. Two real-world datasets (FOS-real and FOS-V) and a synthesized dataset (FOS-syn) are introduced to establish the benchmark dataset. We first elaborate on data collection and categorization of two collected real-world datasets. Then we present how the synthetic test set is constructed with the widely-used CelebA-HQ \cite{celebahq}. Table \ref{tab:overview} presents an overview of the FOS datasets. 
For evaluation, a user study is first conducted to study the effectiveness of the commonly used metrics. 
Based on these conclusions, the following assessment metrics are employed. 1) Traditional IQA metrics, including PSNR, SSIM~\cite{ssim}, LPIPS~\cite{lpips} and FID~\cite{fid}, 2) one new general IQA metric (MANIQA \cite{maniqa}) and one FIQA metric (SER-FIQ \cite{serfiq}).

\subsection{Dataset}
\noindent\textbf{Data collection.}
Our datasets derive from videos on the Internet. Video data instead of image data is collected since face cases from video frames are more diverse and general. This is also a major difference between our datasets and the existing real-world datasets \cite{gfpgan, codeformer}. First, we include YouTube-Faces (YTF)~\cite{ytf_dataset} and YouTube-Celebrities (YTCeleb)~\cite{ytceleb_dataset} datasets as one of our resources. 
Specifically, the YTF dataset contains 3,425 videos of 1,595 different identities, and the average frame number of one video clip is 181.3. The YTCeleb dataset contains 1,910 videos of 47 people with an average of 163.0 frames. 
The frame sizes of these video clips from them are around $[300, 500]$, and 
the proportion of face area is less than 1/8 in most cases. Each frame is associated with unknown degradations, such as compression artifacts, blur, and various noises.
The contents of YTF and YTCeleb datasets are mainly about celebrities talking in show or interview scenes.
To increase the diversity of our test data, we further download 137 videos from YouTube using queries like \textit{old movie}, \textit{aging}, \textit{symphony performance}, and \textit{baseball game} \etc. The average duration of these videos are around 10 minutes. Following VFHQ, we pre-process all the collected raw videos to obtain cropped face video clips. 
The whole process can finally generate 3,316 clips which have the following properties: a) the face area is fully covered and roughly centered; b) the resolution is fixed to $128\times128$; c) the frame length of a single clip is within [50, 1500]; d) a clip only contains one identity. (See more details in supplementary file)

\noindent\textbf{Data categorization.} 
After data collection, we build out test datasets by specifying the collected face images into three categories: 1) \textit{full}: a full face is a front face, and its major facial features (eyes, cheek, nose, mouth, and jaw) are not occluded by other objects; 2) \textit{occluded}: one or more facial features are occluded or truncated; 3) \textit{side}: a side face refers to a face with incomplete facial features (\eg, one eye is hidden) due to a change in head pose.
Figure \ref{fig:overview_fos} presents some examples of three categories. 
We first use Hope-Net~\cite{hopenet} to estimate the head pose and automatically determine a \textit{full} or \textit{side} face.
The head pose estimation network Hope-Net outputs the orientation degrees of one human head regarding three egocentric rotation angles: \textit{yaw}, \textit{pitch} and \textit{roll}~\cite{headpose_survey}. 
Then, we calculate a head pose score by assigning weights to each angle to determine a \textit{side} face.
In this way, a face with a larger head pose movement will be assigned to a greater score. (See details in the supplementary file)
We manually select the occluded subset according to occluded eyes, nose, or mouth from a face image.

\textbf{FOS-syn} consists of pairs of a ground truth face image and the synthesized counterpart. 
The ground-truth images originate from the resized $512\times512$ CelebA-HQ dataset. We first categorize the CelebA-HQ dataset based on the proposed data categorization approach and obtain 1,021, 1,097, and 1,032 images as \textit{full}, \textit{occluded}, and \textit{side} subsets, respectively. 
To generate the synthesized low-quality images, we adopt the commonly used degradation model \cite{gfpgan, restoreformer, vqfr, gcfsr}:
$I^d = \{[(I^h \otimes \textbf{k}_{\sigma})\downarrow_r + \textbf{n}_{\delta}]_{JPEG_q}\}\uparrow_r$, 
where $I^d$, $I^h$ denote the low-quality image and the high-quality counterpart, respectively. 
First, the high-quality image $I^{h}$ is convolved with Gaussian blur kernel $\textbf{k}_{\sigma}$, Then, resampling with scale factor $r$ is performed. Next, additive Gaussian noise $\textbf{n}_{\delta}$ to the resampled image, and the JPEG compression with quality factor $q$ is applied. Finally, the LQ image is resized back to $512\times512$. We follow the widely-used setting in \cite{gfpgan, restoreformer, vqfr} and randomly sample $\sigma$, $r$, $\delta$, and $q$ from $[1, 10]$, $[0.8, 8]$, $[0, 20]$, and $[60, 100]$, respectively. The resampling is based on bilinear interpolation, and the blur kernel size is fixed to 41.

\textbf{FOS-real} is derived from extracted video frames with a stride of 5 from our collected and pre-processed clips. Creating image datasets from video resources can capture more diverse head poses, facial expressions, and gaze directions (see Figure \ref{fig:overview_fos}). 
After data categorization, we obtain 1,633 face images assigned to \textit{full}, 1,136 images assigned to \textit{occluded}, and 1,484 images assigned to \textit{side}.
We exclude video frames with severely truncated/occluded faces or extremely low quality. 

\textbf{FOS-V} is a video test dataset with 3,316 processed face clips from the real world.
This dataset involves heterogeneous scenes such as interviews, sports, nature, and old movies, where the faces are of various ethnicities, across a wide range of ages, and have diverse facial expressions as well as head motions (see Figure \ref{tab:overview} and supplementary file). 

\subsection{Evaluation}
In this section, we unify the evaluation protocol and illustrate the subjective evaluation criteria.
We also specify the used IQA metrics and the proposed new metrics.  

\noindent\textbf{Evaluation protocol.}
For image test sets (\textbf{FOS-syn}, \textbf{FOS-real}), the evaluation is conducted on aligned face images.
Since \textbf{FOS-syn} is already aligned, it can be directly used for quantitative comparison.
For the unaligned \textbf{FOS-real}, we follow \cite{vfhq} to use RetinaNet~\cite{retinanet} for facial landmark detection and OpenCV's \texttt{warpAffine} function for face alignment. This way, we obtain an aligned version of \textbf{FOS-real} for both quantitative assessment and subjective comparison.
The video test sets (\textbf{FOS-V}, VFHQ-Test \cite{vfhq}) are based on a different evaluation protocol. 
In particular, the full-reference metrics are calculated upon the original unaligned video frames, while the no-reference IQA/FIQA metrics are calculated on the aligned video frames for a more specific evaluation on face quality. Therefore, an additional operation -- pasting back to the original LQ video frame is required for full-reference metrics calculation.
Following the impactive works \cite{codeformer, gpen, gfpgan}, we use ParseNet \cite{parsenet} to predict the parsing map of each aligned and restored face image for only pasting the restored face region to the original LQ video frame. This strategy guarantees that only the face area is restored, while the background remains unchanged.

\begin{figure} 
    \vspace{-1em}
    \centering
    \includegraphics[width=.99\linewidth]{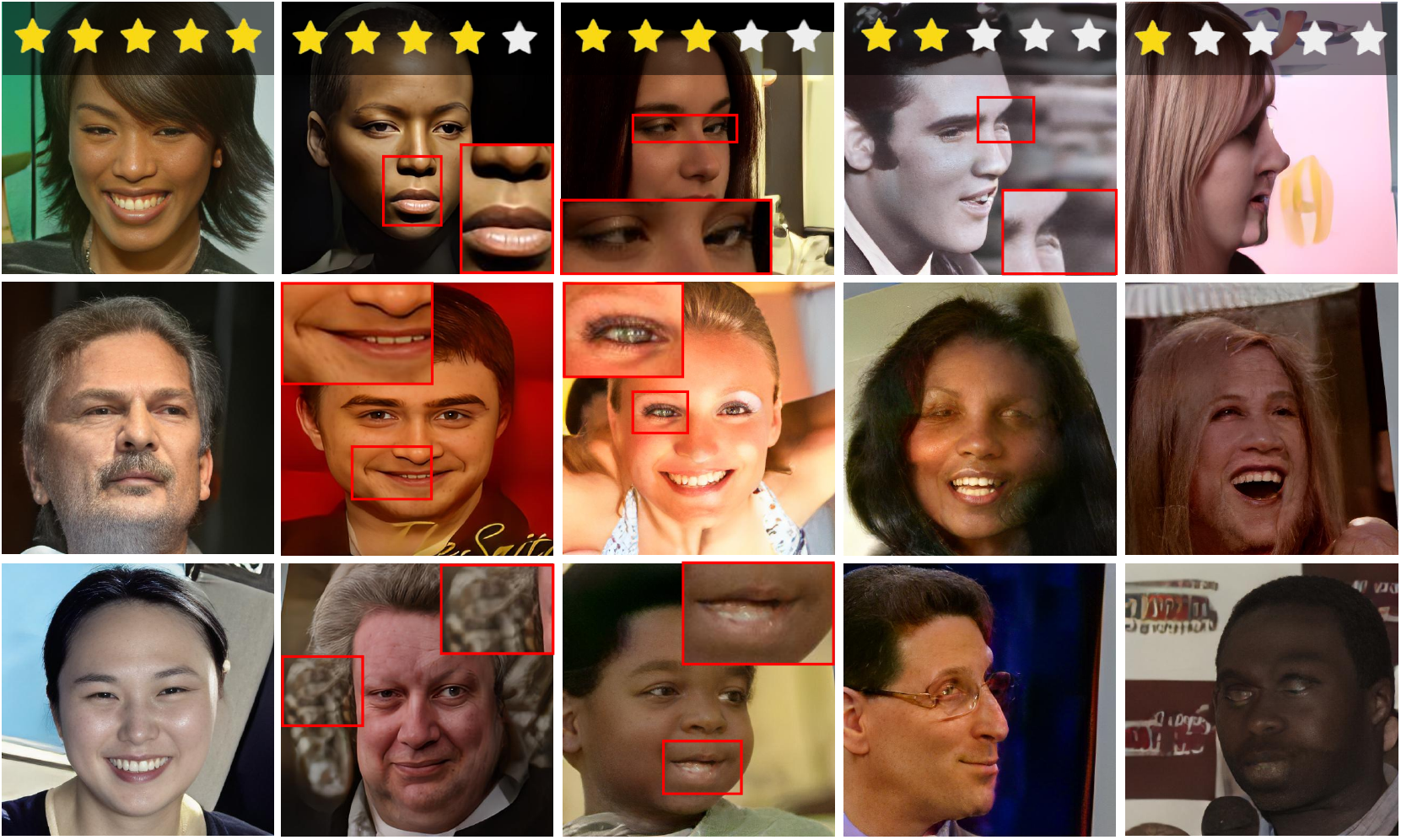}
     \vspace{-0.5em}
    \caption{Visual examples of different scores on our five-point grading system. The number of stars lit corresponds to the score rated. (\textbf{Zoom in for details})}
     \vspace{-1em}
  \label{fig:grade}
\end{figure}

\noindent\textbf{Subjective evaluation criteria.}
For subjective comparison, we conduct a user study to evaluate results on \textbf{FOS-real} and \textbf{FOS-V}. For image comparison on \textbf{FOS-real}, we specify two evaluation dimensions: \textit{fidelity} and \textit{realness}. One is for evaluating the identity preservation of the low-quality face image while the other is for evaluating face generation quality without referring to the LQ input.
The video comparison on \textbf{FOS-V} also has two evaluation dimensions: \textit{stability} (temporal consistency) and \textit{reconstruction performance}. 
For each evaluation dimension, we employ a five-point grading system: 5--Outstanding; 4--Good; 3--Acceptable; 2--Insufficient; 1--Fail.
Note that one-point (Fail) means the restoration performance is far from acceptable, such as exaggerated face distortion regarding \textit{fidelity} and extremely poor image quality regarding \textit{realness}; three-point (Acceptable) indicates the restored result is overall satisfactory but still needs improvement in terms of fine-grained facial attribute reconstruction/generation (\eg, pupil, teeth, \etc.); five-point (Outstanding) means that the restoration result maintains the identity perfectly or is of extraordinary face quality regarding both facial features (\eg, eyes, mouth, \etc.) and texture details (\eg, hair, skin, \etc.). 
Please see Figure \ref{fig:grade} and the supplementary file for visualized illustrations of our five-point grading system.

\noindent\textbf{Evaluation metrics.}
We first adopt several widely acknowledged full-/no-reference metrics (PSNR, SSIM, LPIPS \cite{lpips}, FID) for quantitative measurement. Then, we investigate the effectiveness of five no-reference image quality assessment algorithms (NIQE, BRISQUE~\cite{brisque}, hyperIQA~\cite{hyperiqa}, MUSIQ~\cite{musiq}, MANIQA \cite{maniqa}) and four face image quality assessment algorithms (IFQA \cite{ifqa}, FaceQnet \cite{faceqnet}, SER-FIQ \cite{serfiq}, SDD-FIQA \cite{sddfiq}) by computing the correlation coefficients (PLCC and SROCC) with human opinion scores obtained by user study. It is observed that MANIQA and SER-FIQ are the most relevant with subjective scores (see Figure \ref{fig:mos_image}) among these selected IQA and FIQA algorithms, respectively. Therefore, we use these two metrics on all test datasets.

Moreover, we propose a new Video IDentity Distance (VIDD) metric to evaluate the temporal consistency, \ie. the \textit{stability}, between consecutive video frames. For an input clip consisting of $N$ frames $\{x_1, ..., x_N\}$, a facial feature descriptor $\mathcal{F}$ is used to extract the semantic feature of each input $x_i$. The VIDD score indicates a mean distance of inter-frame features as 
\begin{equation}
    \mathrm{VIDD} = \frac{1}{N} \sum_{i=1}^{N-1} ||\mathcal{F}(x_i), \mathcal{F}(x_{i+1})||_2.
\end{equation}
In particular, we use a pretrained ArcFace~\cite{arcface} model as the descriptor $\mathcal{F}$ to extract discriminative facial features. Referring to Figure \ref{fig:srocc}, the VIDD score demonstrates its significant consistency with the subjective scores, which will be elaborated in detail in section~\ref{section:invest_metrics}.
\section{Experiments}\label{sec:exp}
We evaluate 11 representative state-of-the-art BFR methods
on \textbf{FOS-syn} and \textbf{FOS-real}: GAN prior-based methods (GFP-GAN \cite{gfpgan}, GLEAN \cite{glean}, GPEN \cite{gpen}, PULSE \cite{pulse}), VQ-based methods (VQFR \cite{vqfr}, CodeFormer \cite{codeformer}, RestoreFormer \cite{restoreformer}), reference-based method (DMDNet\footnote{An extension of DFDNet \cite{dfdnet}.} \cite{dmdnet}), and other methods (GCFSR \cite{gcfsr}, PSFRGAN \cite{psfrgan}, HiFaceGAN \cite{hifacegan}). 
Furthermore, we evaluate 4 VSR models trained on VFHQ \cite{vfhq}(EDVR \cite{edvr}, EDVR-GAN, BasicVSR \cite{basicvsr}, BasicVSR-GAN) on our real-world \textbf{FOS-V} dataset and the synthesized VFHQ-Test \cite{vfhq}.  As a comparison, 6 most recent BFR methods (GFP-GAN \cite{gfpgan}, GPEN \cite{gpen}, VQFR \cite{vqfr}, CodeFormer \cite{codeformer}, GCFSR \cite{gcfsr}, RestoreFormer \cite{restoreformer}) are also evaluated on those video datasets in a single-frame processing manner. \footnote{ 
DMDNet is excluded from the comparison since detecting 68 landmarks often fails under complex conditions. }

\subsection{Investigation into The Evaluation Metrics}\label{section:invest_metrics}
The motivation of this section is that we empirically found that the commonly used metrics are inconsistent with the human eye. To better evaluate face restoration results, we first investigate several recently proposed state-of-the-art NR-IQA algorithms (HyperIQA\cite{hyperiqa}, MUSIQ \cite{musiq}, MANIQA \cite{maniqa}), and two traditional NR-IQA algorithms (NIQE \cite{niqe}, BRISQUE \cite{brisque}). 
In addition, we select four recently proposed state-of-the-art FIQA algorithms (IFQA \cite{ifqa}, FaceQnet \cite{faceqnet}, SER-FIQ \cite{serfiq}, SDD-FIQA \cite{sddfiq}) for more extensive exploration. 
Notably, we exclude the commonly used FID since this metric needs to be calculated between data distributions, which may be inaccurate for small-scale test sets. 
\begin{figure}[t]
    \vspace{-1em}
    \centering
    \includegraphics[width=1\linewidth]{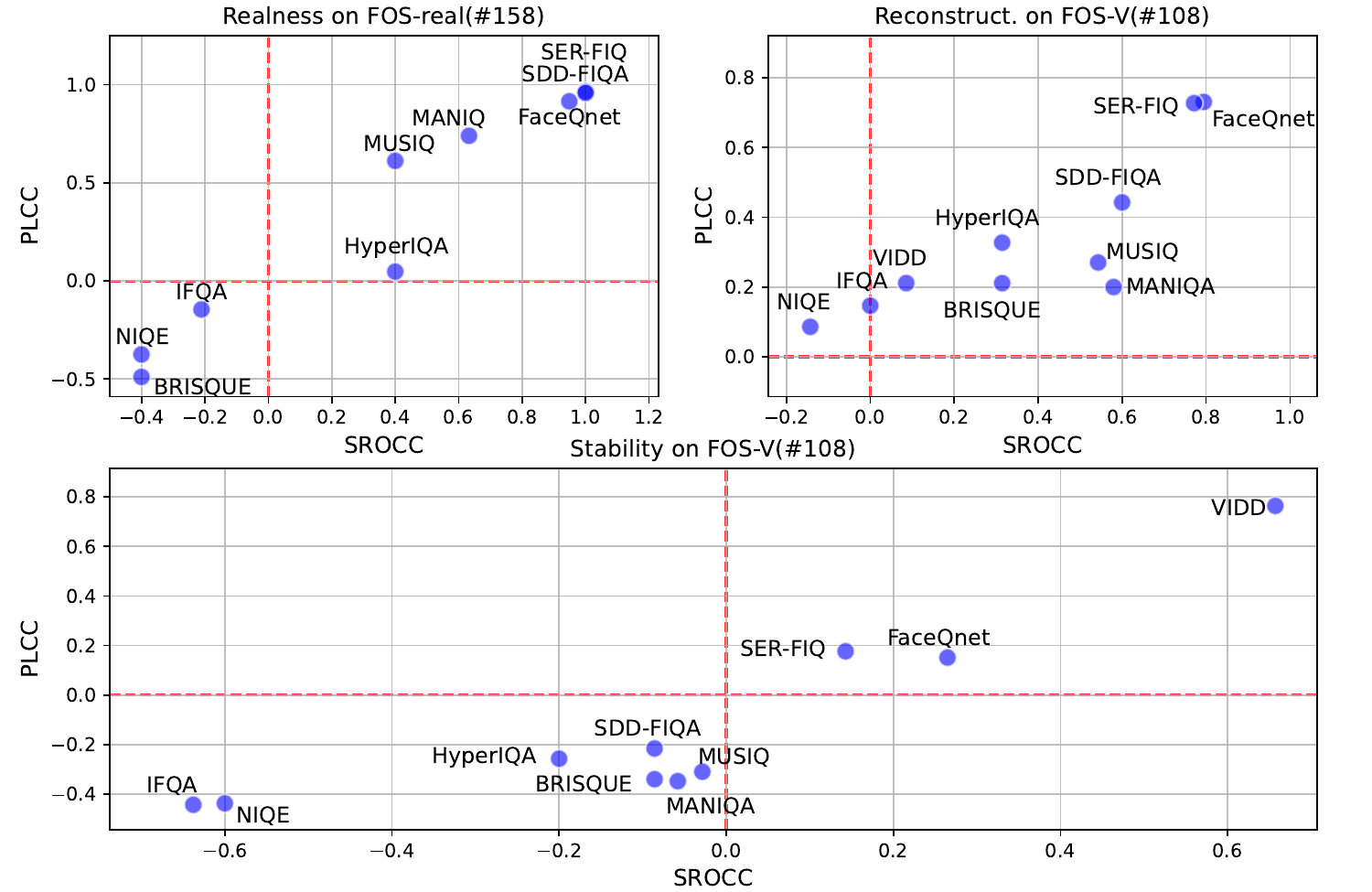}
    \vspace{-1.5em}
    \caption{SROCC v.s PLCC results of 10 IQA/FIQA algorithms and proposed stability evaluation metric VIDD on \textbf{FOS-real}(\#158) regarding \textit{realness} and \textbf{FOS-V}(\#108) regarding \textit{reconstruction performance} and \textit{stability}.}
    \vspace{-1em}
  \label{fig:srocc}
\end{figure}

\begin{table*}[t]
    \vspace{-1em}
    \scriptsize
    \centering
    \setlength\tabcolsep{1.5pt}
    \caption{Comparison of 10 state-of-the-art BFR methods on \textbf{FOS-syn}(\#3150). Results are reported in PSNR, SSIM, LPIPS, SER-FIQ, MANIQA, FID. {\color{red}\textbf{Red}} and {\color{blue}blue} indicate the best and second best results. The top five results are marked as \colorbox{mygray}{gray}.}
    \vspace{-1em}
    \begin{tabular}{@{}c|cccc|cccc|cccc|cccc|cccc|cccc@{}}
    \toprule
              & \multicolumn{4}{c|}{PSNR↑}                                                                                                                                                                                                 & \multicolumn{4}{c|}{SSIM↑}                                                                                                                                                                                                 & \multicolumn{4}{c|}{LPIPS↓}                                                                                                                                                                                                & \multicolumn{4}{c|}{SER-FIQ↑}                                                                                                                                                                                              & \multicolumn{4}{c|}{MANIQA↑}                                                                                                                                                                                               & \multicolumn{4}{c}{FID↓}                                                                                                                                                                                                  \\ \midrule
              & F.                                                   & O.                                                   & S.                                                   & Total                                                & F.                                                   & O.                                                   & S.                                                   & Total                                                & F.                                                   & O.                                                   & S.                                                   & Total                                                & F.                                                   & O.                                                   & S.                                                   & Total                                                & F.                                                   & O.                                                   & S.                                                   & Total                                                & F.                                                   & O.                                                   & S.                                                   & Total                                                \\ \hline
PULSE~\cite{pulse}         & 22.01                                                & 20.56                                                & 21.24                                                & 21.25                                                & 0.623                                                & 0.584                                                & 0.643                                                & 0.616                                                & 0.466                                                & 0.518                                                & 0.513                                                & 0.500                                                & 0.734                                                & \cellcolor[HTML]{F2F2F2}{\color[HTML]{FF0000} 0.700} & \cellcolor[HTML]{F2F2F2}{\color[HTML]{FF0000} 0.661} & \cellcolor[HTML]{F2F2F2}{\color[HTML]{FF0000} 0.698} & 0.590                                                & 0.573                                                & 0.554                                                & 0.572                                                & 88.29                                                & 67.86                                                & 70.66                                                & 75.40                                                \\
PSFR-GAN~\cite{psfrgan}      & 24.30                                                & \cellcolor[HTML]{F2F2F2}23.70                        & \cellcolor[HTML]{F2F2F2}25.15                        & \cellcolor[HTML]{F2F2F2}24.37                        & 0.620                                                & 0.596                                                & 0.669                                                & 0.628                                                & 0.409                                                & 0.444                                                & 0.440                                                & 0.431                                                & 0.515                                                & 0.413                                                & 0.365                                                & 0.430                                                & 0.626                                                & 0.612                                                & 0.582                                                & 0.606                                                & \cellcolor[HTML]{F2F2F2}{\color[HTML]{1552D1} 68.77} & \cellcolor[HTML]{F2F2F2}{\color[HTML]{1552D1} 58.86} & \cellcolor[HTML]{F2F2F2}63.38                        & \cellcolor[HTML]{F2F2F2}63.56                        \\
HiFaceGAN~\cite{hifacegan}     & \cellcolor[HTML]{F2F2F2}24.75                        & 23.64                                                & 24.71                                                & 24.35                                                & 0.622                                                & 0.592                                                & 0.644                                                & 0.618                                                & 0.418                                                & 0.456                                                & 0.460                                                & 0.445                                                & 0.766                                                & 0.616                                                & 0.531                                                & 0.637                                                & 0.595                                                & 0.553                                                & 0.530                                                & 0.559                                                & \cellcolor[HTML]{F2F2F2}71.90                        & \cellcolor[HTML]{F2F2F2}62.59                        & 72.12                                                & 68.73                                                \\
GPEN~\cite{gpen}          & 24.60                                                & 23.41                                                & 24.75                                                & 24.24                                                & \cellcolor[HTML]{F2F2F2}0.664                        & \cellcolor[HTML]{F2F2F2}0.635                        & \cellcolor[HTML]{F2F2F2}0.707                        & \cellcolor[HTML]{F2F2F2}0.668                        & 0.394                                                & 0.402                                                & 0.388                                                & 0.395                                                & 0.768                                                & 0.629                                                & 0.556                                                & 0.650                                                & \cellcolor[HTML]{F2F2F2}{\color[HTML]{FF0000} 0.695} & \cellcolor[HTML]{F2F2F2}{\color[HTML]{FF0000} 0.695} & \cellcolor[HTML]{F2F2F2}{\color[HTML]{FF0000} 0.674} & \cellcolor[HTML]{F2F2F2}{\color[HTML]{FF0000} 0.688} & \cellcolor[HTML]{F2F2F2}69.39                        & 63.67                                                & 66.95                                                & \cellcolor[HTML]{F2F2F2}66.60                        \\
GCFSR~\cite{gcfsr}         & \cellcolor[HTML]{F2F2F2}{\color[HTML]{FF0000} 26.31} & \cellcolor[HTML]{F2F2F2}{\color[HTML]{0B5FD1} 25.11} & \cellcolor[HTML]{F2F2F2}{\color[HTML]{FF0000} 26.43} & \cellcolor[HTML]{F2F2F2}{\color[HTML]{FF0000} 25.93} & \cellcolor[HTML]{F2F2F2}{\color[HTML]{FF0000} 0.699} & \cellcolor[HTML]{F2F2F2}{\color[HTML]{0B5FD1} 0.677} & \cellcolor[HTML]{F2F2F2}{\color[HTML]{FF0000} 0.734} & \cellcolor[HTML]{F2F2F2}{\color[HTML]{FF0000} 0.703} & \cellcolor[HTML]{F2F2F2}{\color[HTML]{FF0000} 0.311} & \cellcolor[HTML]{F2F2F2}{\color[HTML]{FF0000} 0.353} & \cellcolor[HTML]{F2F2F2}{\color[HTML]{FF0000} 0.354} & \cellcolor[HTML]{F2F2F2}{\color[HTML]{FF0000} 0.340} & \cellcolor[HTML]{F2F2F2}0.772                        & 0.654                                                & \cellcolor[HTML]{F2F2F2}0.579                        & 0.668                                                & \cellcolor[HTML]{F2F2F2}0.656                        & \cellcolor[HTML]{F2F2F2}0.640                        & \cellcolor[HTML]{F2F2F2}0.622                        & \cellcolor[HTML]{F2F2F2}0.639                        & 79.71                                                & 64.36                                                & \cellcolor[HTML]{F2F2F2}65.31                        & 69.65                                                \\
DMDNet~\cite{dmdnet}        & \cellcolor[HTML]{F2F2F2}{\color[HTML]{0B5FD1} 25.67} & \cellcolor[HTML]{F2F2F2}24.48                        & \cellcolor[HTML]{F2F2F2}{\color[HTML]{0B5FD1} 25.79} & \cellcolor[HTML]{F2F2F2}{\color[HTML]{0B5FD1} 25.30} & \cellcolor[HTML]{F2F2F2}{\color[HTML]{0B5FD1} 0.681} & \cellcolor[HTML]{F2F2F2}0.656                        & \cellcolor[HTML]{F2F2F2}{\color[HTML]{0B5FD1} 0.719} & \cellcolor[HTML]{F2F2F2}{\color[HTML]{0B5FD1} 0.685} & 0.355                                                & 0.398                                                & 0.389                                                & 0.381                                                & \cellcolor[HTML]{F2F2F2}0.769                        & 0.637                                                & 0.555                                                & 0.653                                                & 0.622                                                & 0.586                                                & 0.558                                                & 0.588                                                & \cellcolor[HTML]{F2F2F2}{\color[HTML]{FF0000} 67.61} & \cellcolor[HTML]{F2F2F2}60.91                        & \cellcolor[HTML]{F2F2F2}{\color[HTML]{1552D1} 61.91} & \cellcolor[HTML]{F2F2F2}{\color[HTML]{1552D1} 63.41} \\
GFP-GAN~\cite{gfpgan}       & \cellcolor[HTML]{F2F2F2}25.27                        & \cellcolor[HTML]{F2F2F2}23.95                        & \cellcolor[HTML]{F2F2F2}25.13                        & \cellcolor[HTML]{F2F2F2}24.76                        & \cellcolor[HTML]{F2F2F2}0.673                        & \cellcolor[HTML]{F2F2F2}0.642                        & \cellcolor[HTML]{F2F2F2}0.702                        & \cellcolor[HTML]{F2F2F2}0.672                        & \cellcolor[HTML]{F2F2F2}0.331                        & \cellcolor[HTML]{F2F2F2}0.381                        & \cellcolor[HTML]{F2F2F2}0.385                        & \cellcolor[HTML]{F2F2F2}0.366                        & \cellcolor[HTML]{F2F2F2}0.771                        & \cellcolor[HTML]{F2F2F2}0.662                        & \cellcolor[HTML]{F2F2F2}0.589                        & \cellcolor[HTML]{F2F2F2}0.673                        & \cellcolor[HTML]{F2F2F2}0.664                        & \cellcolor[HTML]{F2F2F2}0.660                        & \cellcolor[HTML]{F2F2F2}0.641                        & \cellcolor[HTML]{F2F2F2}0.655                        & 78.06                                                & 62.75                                                & \cellcolor[HTML]{F2F2F2}63.39                        & 67.92                                                \\
VQFR~\cite{vqfr}          & 23.95                                                & 22.49                                                & 23.69                                                & 23.36                                                & 0.650                                                & 0.618                                                & 0.685                                                & 0.651                                                & \cellcolor[HTML]{F2F2F2}0.328                        & \cellcolor[HTML]{F2F2F2}0.373                        & \cellcolor[HTML]{F2F2F2}0.375                        & \cellcolor[HTML]{F2F2F2}0.359                        & 0.765                                                & \cellcolor[HTML]{F2F2F2}0.664                        & \cellcolor[HTML]{F2F2F2}0.587                        & \cellcolor[HTML]{F2F2F2}0.672                        & \cellcolor[HTML]{F2F2F2}0.654                        & \cellcolor[HTML]{F2F2F2}0.644                        & \cellcolor[HTML]{F2F2F2}0.624                        & \cellcolor[HTML]{F2F2F2}0.641                        & 75.90                                                & \cellcolor[HTML]{F2F2F2}61.68                        & 66.26                                                & \cellcolor[HTML]{F2F2F2}67.79                        \\
RestoreFormer~\cite{restoreformer} & 24.74                                                & 23.41                                                & 24.68                                                & 24.26                                                & 0.636                                                & 0.609                                                & 0.671                                                & 0.638                                                & \cellcolor[HTML]{F2F2F2}0.328                        & \cellcolor[HTML]{F2F2F2}0.373                        & \cellcolor[HTML]{F2F2F2}0.381                        & \cellcolor[HTML]{F2F2F2}0.361                        & \cellcolor[HTML]{F2F2F2}{\color[HTML]{0B5FD1} 0.781} & \cellcolor[HTML]{F2F2F2}0.656                        & 0.576                                                & \cellcolor[HTML]{F2F2F2}0.670                        & 0.648                                                & 0.634                                                & 0.613                                                & 0.631                                                & \cellcolor[HTML]{F2F2F2}72.09                        & \cellcolor[HTML]{F2F2F2}{\color[HTML]{FF0000} 56.92} & \cellcolor[HTML]{F2F2F2}{\color[HTML]{FF0000} 56.81} & \cellcolor[HTML]{F2F2F2}{\color[HTML]{FF0000} 61.80} \\
CodeFormer~\cite{codeformer}    & \cellcolor[HTML]{F2F2F2}25.29                        & \cellcolor[HTML]{F2F2F2}{\color[HTML]{FF0000} 25.17} & \cellcolor[HTML]{F2F2F2}25.17                        & \cellcolor[HTML]{F2F2F2}25.21                        & \cellcolor[HTML]{F2F2F2}0.661                        & \cellcolor[HTML]{F2F2F2}{\color[HTML]{FF0000} 0.692} & \cellcolor[HTML]{F2F2F2}0.692                        & \cellcolor[HTML]{F2F2F2}0.682                        & \cellcolor[HTML]{F2F2F2}{\color[HTML]{0B5FD1} 0.317} & \cellcolor[HTML]{F2F2F2}{\color[HTML]{0B5FD1} 0.358} & \cellcolor[HTML]{F2F2F2}{\color[HTML]{0B5FD1} 0.358} & \cellcolor[HTML]{F2F2F2}{\color[HTML]{0B5FD1} 0.345} & \cellcolor[HTML]{F2F2F2}{\color[HTML]{FF0000} 0.783} & \cellcolor[HTML]{F2F2F2}{\color[HTML]{0B5FD1} 0.680} & \cellcolor[HTML]{F2F2F2}{\color[HTML]{0B5FD1} 0.603} & \cellcolor[HTML]{F2F2F2}{\color[HTML]{0B5FD1} 0.688} & \cellcolor[HTML]{F2F2F2}{\color[HTML]{0B5FD1} 0.666} & \cellcolor[HTML]{F2F2F2}{\color[HTML]{0B5FD1} 0.666} & \cellcolor[HTML]{F2F2F2}{\color[HTML]{0B5FD1} 0.645} & \cellcolor[HTML]{F2F2F2}{\color[HTML]{0B5FD1} 0.659} & 80.90                                                & 65.02                                                & 68.78                                                & 71.40                                                \\ \bottomrule
    \end{tabular}
    \label{tab:fos-syn}
    \vspace{-0.5em}
\end{table*}

\begin{table*}[htbp]
    \centering
    \caption{Comparison of 11 state-of-the-art BFR methods on \textbf{FOS-real}(\#4253). Results are reported in SER-FIQ, MANIQA, and FID.  {\color{red}\textbf{Red}} and {\color{blue}blue} indicate the best and second best performance. The top five results are marked as \colorbox{mygray}{gray}.}
    \vspace{-1em}
    \footnotesize
    \setlength{\baselineskip}{22pt}
    \setlength\tabcolsep{6pt}
    \begin{tabular}{@{}c|cccc|cccc|cccc@{}}
    \toprule
              & \multicolumn{4}{c|}{SER-FIQ↑}                                                                                                                                                                                              & \multicolumn{4}{c|}{MANIQA↑}                                                                                                                                                                                               & \multicolumn{4}{c}{FID↓}                                                                                                                                                                                                  \\ \midrule
              & \multicolumn{1}{c}{F.}                               & \multicolumn{1}{c}{O.}                               & \multicolumn{1}{c}{S.}                               & Total                                                & F.                                                   & O.                                                   & S.                                                   & Total                                                & F.                                                   & O.                                                   & S.                                                   & Total                                                \\
PULSE~\cite{pulse}         & \cellcolor[HTML]{F2F2F2}{\color[HTML]{FF0000} 0.710} & \cellcolor[HTML]{F2F2F2}{\color[HTML]{FF0000} 0.674} & \cellcolor[HTML]{F2F2F2}{\color[HTML]{FF0000} 0.638} & \cellcolor[HTML]{F2F2F2}{\color[HTML]{FF0000} 0.675} & 0.565                                                & 0.549                                                & 0.542                                                & 0.553                                                & 58.90                                                & \cellcolor[HTML]{F2F2F2}{\color[HTML]{FF0000} 61.28} & \cellcolor[HTML]{F2F2F2}{\color[HTML]{FF0000} 61.16} & 60.32                                                \\
PSFR-GAN~\cite{psfrgan}      & 0.664                                                & 0.511                                                & 0.379                                                & 0.523                                                & 0.579                                                & 0.550                                                & 0.532                                                & 0.555                                                & 46.47                                                & \cellcolor[HTML]{F2F2F2}64.93                        & 75.74                                                & 61.61                                                \\
HiFaceGAN~\cite{hifacegan}     & 0.631                                                & 0.469                                                & 0.313                                                & 0.477                                                & 0.502                                                & 0.463                                                & 0.427                                                & 0.465                                                & 63.46                                                & 94.00                                                & 117.88                                               & 90.61                                                \\
GLEAN~\cite{glean}         & \cellcolor[HTML]{F2F2F2}0.685                        & 0.511                                                & 0.387                                                & 0.534                                                & 0.591                                                & 0.567                                                & 0.529                                                & 0.563                                                & 52.07                                                & 69.79                                                & 79.95                                                & 66.53                                                \\
GPEN~\cite{gpen}          & 0.679                                                & 0.519                                                & \cellcolor[HTML]{F2F2F2}0.406                        & 0.541                                                & \cellcolor[HTML]{F2F2F2}{\color[HTML]{FF0000} 0.672} & \cellcolor[HTML]{F2F2F2}{\color[HTML]{FF0000} 0.661} & \cellcolor[HTML]{F2F2F2}{\color[HTML]{FF0000} 0.641} & \cellcolor[HTML]{F2F2F2}{\color[HTML]{FF0000} 0.658} & 57.40                                                & 76.37                                                & 82.45                                                & 71.21                                                \\
GCFSR~\cite{gcfsr}         & 0.675                                                & 0.517                                                & 0.391                                                & 0.534                                                & \cellcolor[HTML]{F2F2F2}0.609                        & \cellcolor[HTML]{F2F2F2}0.582                        & \cellcolor[HTML]{F2F2F2}0.560                        & \cellcolor[HTML]{F2F2F2}0.585                        & 46.77                                                & 69.79                                                & 68.65                                                & 60.55                                                \\
DMDNet~\cite{dmdnet}        & 0.673                                                & 0.507                                                & 0.373                                                & 0.524                                                & 0.578                                                & 0.543                                                & 0.509                                                & 0.545                                                & \cellcolor[HTML]{F2F2F2}{\color[HTML]{1552D1} 41.25} & 65.76                                                & 70.68                                                & \cellcolor[HTML]{F2F2F2}58.07                        \\
GFP-GAN~\cite{gfpgan}       & \cellcolor[HTML]{F2F2F2}0.687                        & \cellcolor[HTML]{F2F2F2}0.532                        & \cellcolor[HTML]{F2F2F2}0.419                        & \cellcolor[HTML]{F2F2F2}0.552                        & \cellcolor[HTML]{F2F2F2}{\color[HTML]{0B5FD1} 0.638} & \cellcolor[HTML]{F2F2F2}{\color[HTML]{0B5FD1} 0.622} & \cellcolor[HTML]{F2F2F2}0.599                        & \cellcolor[HTML]{F2F2F2}{\color[HTML]{0B5FD1} 0.620} & \cellcolor[HTML]{F2F2F2}44.27                        & 65.50                                                & \cellcolor[HTML]{F2F2F2}64.44                        & \cellcolor[HTML]{F2F2F2}56.98                        \\
VQFR~\cite{vqfr}          & \cellcolor[HTML]{F2F2F2}0.695                        & \cellcolor[HTML]{F2F2F2}0.533                        & \cellcolor[HTML]{F2F2F2}0.425                        & \cellcolor[HTML]{F2F2F2}0.558                        & \cellcolor[HTML]{F2F2F2}0.614                        & \cellcolor[HTML]{F2F2F2}0.593                        & \cellcolor[HTML]{F2F2F2}0.567                        & \cellcolor[HTML]{F2F2F2}0.592                        & \cellcolor[HTML]{F2F2F2}42.99                        & \cellcolor[HTML]{F2F2F2}65.25                        & \cellcolor[HTML]{F2F2F2}65.33                        & \cellcolor[HTML]{F2F2F2}56.73                        \\
RestoreFormer~\cite{restoreformer} & 0.683                                                & \cellcolor[HTML]{F2F2F2}0.522                        & 0.402                                                & \cellcolor[HTML]{F2F2F2}0.542                        & 0.596                                                & 0.576                                                & 0.541                                                & 0.572                                                & \cellcolor[HTML]{F2F2F2}{\color[HTML]{FF0000} 39.85} & \cellcolor[HTML]{F2F2F2}64.98                        & \cellcolor[HTML]{F2F2F2}66.53                        & \cellcolor[HTML]{F2F2F2}{\color[HTML]{FF0000} 55.87} \\
CodeFormer~\cite{codeformer}    & \cellcolor[HTML]{F2F2F2}{\color[HTML]{0B5FD1} 0.709} & \cellcolor[HTML]{F2F2F2}{\color[HTML]{0B5FD1} 0.557} & \cellcolor[HTML]{F2F2F2}{\color[HTML]{0B5FD1} 0.444} & \cellcolor[HTML]{F2F2F2}{\color[HTML]{0B5FD1} 0.576} & \cellcolor[HTML]{F2F2F2}0.634                        & \cellcolor[HTML]{F2F2F2}0.621                        & \cellcolor[HTML]{F2F2F2}{\color[HTML]{0B5FD1} 0.599} & \cellcolor[HTML]{F2F2F2}0.618                        & \cellcolor[HTML]{F2F2F2}43.98                        & \cellcolor[HTML]{F2F2F2}{\color[HTML]{1552D1} 63.09} & \cellcolor[HTML]{F2F2F2}{\color[HTML]{1552D1} 63.96} & \cellcolor[HTML]{F2F2F2}{\color[HTML]{1552D1} 56.05} \\ \bottomrule
\end{tabular}
    \label{tab:fos-real}
    \vspace{-1em}
\end{table*}

We select 158 image samples from \textbf{FOS-real} as \textbf{FOS-real} (\#158), and 108 video samples as \textbf{FOS-V} (\#108). Methods on these two selected datasets are used for user rating. Based on the obtained subjective scores regarding the \textit{realness} evaluation dimension, we calculate their correlation coefficients with the results of these selected IQA/FIQA algorithms. The SROCC v.s PLCC results are illustrated in Figure \ref{fig:srocc}. Three FIQA algorithms, SER-FIQ, SDD-FIQA, and FaceQnet exhibit tremendously strong correlations with human ratings. For IQA algorithms, MANIQA and MUSIQ show good consistency. However, the commonly used NIQE presents a negative correlation. 
Furthermore, we provide the SROCC v.s. PLCC results on \textbf{FOS-V}(\#108) in terms of \textit{reconstruction performance}. SER-FIQ is still fairly aligned with subjective scores, while SDD-FIQA, MANIQA, and MUSIQ show an obvious performance drop. In addition, although we exclude FID from the investigation, it yields noticeable relevance with SER-FIQ scores (see Table \ref{tab:fos-real}). 

\begin{table}[]
    \scriptsize
    \setlength\tabcolsep{1pt}
    \centering
    \caption{Comparison of 6 BFR and 4 VSR methods on \textbf{VFHQ-Test}(\#50). {\color{red}\textbf{Red}} and {\color{blue}blue} indicate the best and second best results. Top five results are marked as \colorbox{mygray}{gray}.}
    \vspace{-1em}
    \begin{tabular}{@{}cccccccc@{}}
\toprule
              & \multicolumn{1}{l}{PSNR↑}                             & \multicolumn{1}{l}{SSIM↑}                            & \multicolumn{1}{l}{LPIPS↓}                           & \multicolumn{1}{l}{SER-FIQ↑}                         & MANIQA↑                                              & FID↓                                                 & VIDD↓                                               \\ \midrule
GPEN~\cite{gpen}          & 26.232                                                & 0.766                                                & 0.368                                                & \cellcolor[HTML]{F2F2F2}0.674                        & \cellcolor[HTML]{F2F2F2}{\color[HTML]{FF0000} 0.580} & \cellcolor[HTML]{F2F2F2}89.85                        & \cellcolor[HTML]{F2F2F2}0.41                        \\
GCFSR~\cite{gcfsr}         & \cellcolor[HTML]{F2F2F2}26.747                        & \cellcolor[HTML]{F2F2F2}0.783                        & \cellcolor[HTML]{F2F2F2}0.355                        & 0.662                                                & 0.533                                                & 98.63                                                & \cellcolor[HTML]{F2F2F2}0.42                        \\
GFP-GAN~\cite{gfpgan}       & \cellcolor[HTML]{F2F2F2}26.547                        & \cellcolor[HTML]{F2F2F2}0.776                        & 0.360                                                & 0.670                                                & \cellcolor[HTML]{F2F2F2}0.551                        & 95.08                                                & 0.42                                                \\
VQFR~\cite{vqfr}          & 25.645                                                & 0.760                                                & 0.365                                                & \cellcolor[HTML]{F2F2F2}{\color[HTML]{1552D1} 0.677} & 0.547                                                & \cellcolor[HTML]{F2F2F2}91.98                        & 0.58                                                \\
RestoreFormer~\cite{restoreformer} & 26.094                                                & 0.755                                                & 0.381                                                & 0.649                                                & \cellcolor[HTML]{F2F2F2}0.552                        & \cellcolor[HTML]{F2F2F2}{\color[HTML]{FF0000} 87.26} & 0.46                                                \\
CodeFormer~\cite{codeformer}    & \cellcolor[HTML]{F2F2F2}26.420                        & \cellcolor[HTML]{F2F2F2}0.771                        & 0.360                                                & \cellcolor[HTML]{F2F2F2}{\color[HTML]{FF0000} 0.699} & \cellcolor[HTML]{F2F2F2}0.552                        & 97.16                                                & 0.46                                                \\
BasicVSR~\cite{basicvsr}      & \cellcolor[HTML]{F2F2F2}{\color[HTML]{FF0000} 29.353} & \cellcolor[HTML]{F2F2F2}{\color[HTML]{FF0000} 0.848} & \cellcolor[HTML]{F2F2F2}0.319                        & 0.671                                                & 0.326                                                & 131.45                                               & \cellcolor[HTML]{F2F2F2}{\color[HTML]{FF0000} 0.31} \\
EDVR~\cite{edvr}          & \cellcolor[HTML]{F2F2F2}{\color[HTML]{1552D1} 29.269} & \cellcolor[HTML]{F2F2F2}{\color[HTML]{1552D1} 0.846} & \cellcolor[HTML]{F2F2F2}0.322                        & \cellcolor[HTML]{F2F2F2}0.677                        & 0.326                                                & 131.02                                               & \cellcolor[HTML]{F2F2F2}{\color[HTML]{1552D1} 0.33} \\
EDVR-GAN~\cite{edvr}      & 26.384                                                & 0.766                                                & \cellcolor[HTML]{F2F2F2}{\color[HTML]{FF0000} 0.304} & \cellcolor[HTML]{F2F2F2}0.674                        & \cellcolor[HTML]{F2F2F2}{\color[HTML]{1552D1} 0.555} & \cellcolor[HTML]{F2F2F2}89.14                        & \cellcolor[HTML]{F2F2F2}0.39                        \\
BasicVSR-GAN~\cite{basicvsr}  & 25.820                                                & 0.759                                                & \cellcolor[HTML]{F2F2F2}{\color[HTML]{1552D1} 0.319} & 0.651                                                & 0.550                                                & \cellcolor[HTML]{F2F2F2}{\color[HTML]{1552D1} 87.45} & 0.44                                                \\ \bottomrule
\vspace{-2em}
\end{tabular}
    \label{tab:vfhq_performance}
    \vspace{-1em}
\end{table}

Based on these observations, we adopt the best-performing FIQA and IQA algorithms, SER-FIQ, and MANIQA to evaluate all test sets.  
More importantly, we highly recommend that readers focus more on SER-FIQ scores than MANIQA and FID in the paper.
To demonstrate the effectiveness of our proposed VIDD, we also compute the correlation coefficients concerning the obtained subjective scores on \textbf{FOS-V}(\#108) according to \textit{stability}. In the bottom Figure \ref{fig:srocc}, the proposed VIDD yields 0.763 on SROCC and 0.657 on PLCC, which stands out among all explored metrics.

\subsection{Benchmark Results on Image Datasets.}
\noindent\textbf{FOS-syn.} 
We first compare the results of 10 blind face restoration methods\footnote{We exclude GLEAN for comparison as its blind version involves training on CelebA-HQ dataset.} on the synthesized FOS-syn dataset.
As seen from Table \ref{tab:fos-syn}, GCFSR, CodeFormer, and GFP-GAN could achieve more satisfactory results in terms of all full-reference metrics (PSNR, SSIM, and LPIPS), indicating their great identity preservation and perceptual quality. 
Regarding the no-reference metrics, we observe that PULSE obtains the best results in SER-FIQ, but performs the worst in PSNR, SSIM, and LPIPS. This manifests that PULSE generates high-quality face images without considering maintaining the identity information. 
In comparison, CodeFormer, RestoreFormer, VQFR, GFP-GAN gain good results on both SER-FIQ and LPIPS. 
In summary, on the synthesized dataset, GCFSR and DMDNet \cite{dmdnet} show superior restoration performance, while RestoreFormer and CodeFormer exhibit better generation ability. No method can surpass the others on all metrics and datasets.

\noindent\textbf{FOS-real.}
We evaluate 11 BFR methods on our real-world image dataset. The qualitative comparison can be found in Figure \ref{fig:fos_performance_overview} (see more results in supplementary file). Three no-reference metrics FID, MANIQA \cite{maniqa}, and SER-FIQ \cite{serfiq} are adopted for quantitative comparison. Table \ref{tab:fos-real} reports the performance comparison. 
By excluding the results of PULSE, we find that CodeFormer, VQFR, GFP-GAN, and RestoreFormer yield superior performance regarding the FIQA metric SER-FIQ. 
Regarding the IQA metric MANIQA, the top 5 methods are GPEN, GFP-GAN, CodeFormer, VQFR, and GCFSR.
By carefully examining the FID results, we observe that CodeFormer, RestoreFormer, VQFR, and GFP-GAN attain more satisfactory results than other methods. 
Notably, all methods perform worse on both \textit{occluded} and \textit{side} subsets, especially regarding FID and SER-FIQ. However, MANIQA shows less significant differences among the three subsets since it is a general IQA metric. 

\begin{table}[]
    \vspace{-1em}
    \footnotesize
    \setlength\tabcolsep{1pt}
    \centering
    \caption{Comparison of 6 BFR and 4 VSR methods on \textbf{FOS-V}(\#3316). {\color{red}\textbf{Red}} and {\color{blue}blue} indicate the best and second best performance. Top five results are marked as \colorbox{mygray}{gray}.}
    \vspace{-0.5em}
    \begin{tabular}{@{}ccccc@{}}
\toprule
              & \multicolumn{1}{l}{SER-FIQ↑}                         & MANIQA↑                                              & FID↓                                                 & VIDD↓                                               \\ \midrule
GPEN~\cite{gpen}          & \cellcolor[HTML]{F2F2F2}0.596                        & \cellcolor[HTML]{F2F2F2}{\color[HTML]{FF0000} 0.639} & \cellcolor[HTML]{F2F2F2}{\color[HTML]{FF0000} 79.21} & 0.51                                                \\
GCFSR~\cite{gcfsr}         & \cellcolor[HTML]{F2F2F2}0.572                        & 0.473                                                & 98.53                                                & \cellcolor[HTML]{F2F2F2}0.48                        \\
GFP-GAN~\cite{gfpgan}       & \cellcolor[HTML]{F2F2F2}{\color[HTML]{1552D1} 0.601} & \cellcolor[HTML]{F2F2F2}0.515                        & 96.50                                                & \cellcolor[HTML]{F2F2F2}0.48                        \\
VQFR~\cite{vqfr}          & \cellcolor[HTML]{F2F2F2}0.596                        & \cellcolor[HTML]{F2F2F2}0.514                        & \cellcolor[HTML]{F2F2F2}85.60                        & 0.62                                                \\
RestoreFormer~\cite{restoreformer} & 0.556                                                & \cellcolor[HTML]{F2F2F2}{\color[HTML]{1552D1} 0.524} & \cellcolor[HTML]{F2F2F2}86.82                        & 0.50                                                \\
CodeFormer~\cite{codeformer}    & \cellcolor[HTML]{F2F2F2}{\color[HTML]{FF0000} 0.616} & \cellcolor[HTML]{F2F2F2}0.520                        & 98.85                                                & 0.50                                                \\
BasicVSR~\cite{basicvsr}      & 0.567                                                & 0.300                                                & 123.26                                               & \cellcolor[HTML]{F2F2F2}{\color[HTML]{FF0000} 0.36} \\
EDVR~\cite{edvr}          & 0.567                                                & 0.302                                                & 124.20                                               & \cellcolor[HTML]{F2F2F2}{\color[HTML]{1552D1} 0.38} \\
EDVR-GAN~\cite{edvr}      & 0.562                                                & 0.499                                                & \cellcolor[HTML]{F2F2F2}84.84                        & \cellcolor[HTML]{F2F2F2}0.44                        \\
BasicVSR-GAN~\cite{basicvsr}  & 0.520                                                & 0.505                                                & \cellcolor[HTML]{F2F2F2}{\color[HTML]{1552D1} 83.19} & 0.52                                                \\ \bottomrule
\end{tabular}
    \vspace{-1.5em}
    \label{tab:fos_v_performance}
\end{table}

\subsection{Benchmark Results on Video Datasets.}

\noindent\textbf{FOS-V.} 
We also conduct a performance comparison among BFR and VSR methods on our \text{FOS-V} dataset (See Table \ref{tab:fos_v_performance}). The VSR methods achieve the best performance in VIDD, showing high stability on the video dataset. For IQA metrics, the BFR methods achieve superior overall performance compared to VSR methods. However, they fall behind in maintaining video restoration stability referring to VIDD. Among BFR methods, GPEN obtains excellent performance in both FID and the general metric MANIQA. CodeFormer demonstrates its superiority in real-world scenes with a significantly higher SER-FIQ score, but its stability falls out of the top five. GFP-GAN ranks second in the SER-FIQ score and achieves good performance regarding VIDD, indicating its excellent balance on \textit{reconstruction performance} and \textit{stability}.

We additionally leverage a synthesized video test set, \textbf{VFHQ-Test}, to our evaluation. 
We compare 6 state-of-the-art BFR and 4 VSR ($\times4$) methods on the synthesized VFHQ-Test in Table \ref{tab:vfhq_performance}.
BasicVSR obtains the best result in PSNR and SSIM, while EDVR ranks second. As for LPIPS, EDVR-GAN and BasicVSR-GAN achieve first and second place, respectively. This implies that GAN training improves perceptual quality.
For inter-frame stability (VIDD), BasicVSR and EDVR yield significantly better results than others due to the multi-frame and MSE-oriented training strategies.
Regarding the no-reference IQA/FIQA metrics (SER-FIQ, MANIQA, and FID), CodeFormer, GPEN, and RestoreFormer achieve superior results over other methods.


\begin{figure*}[htbp]
    \centering
    \vspace{-1.3em}
    \includegraphics[width=1\linewidth]{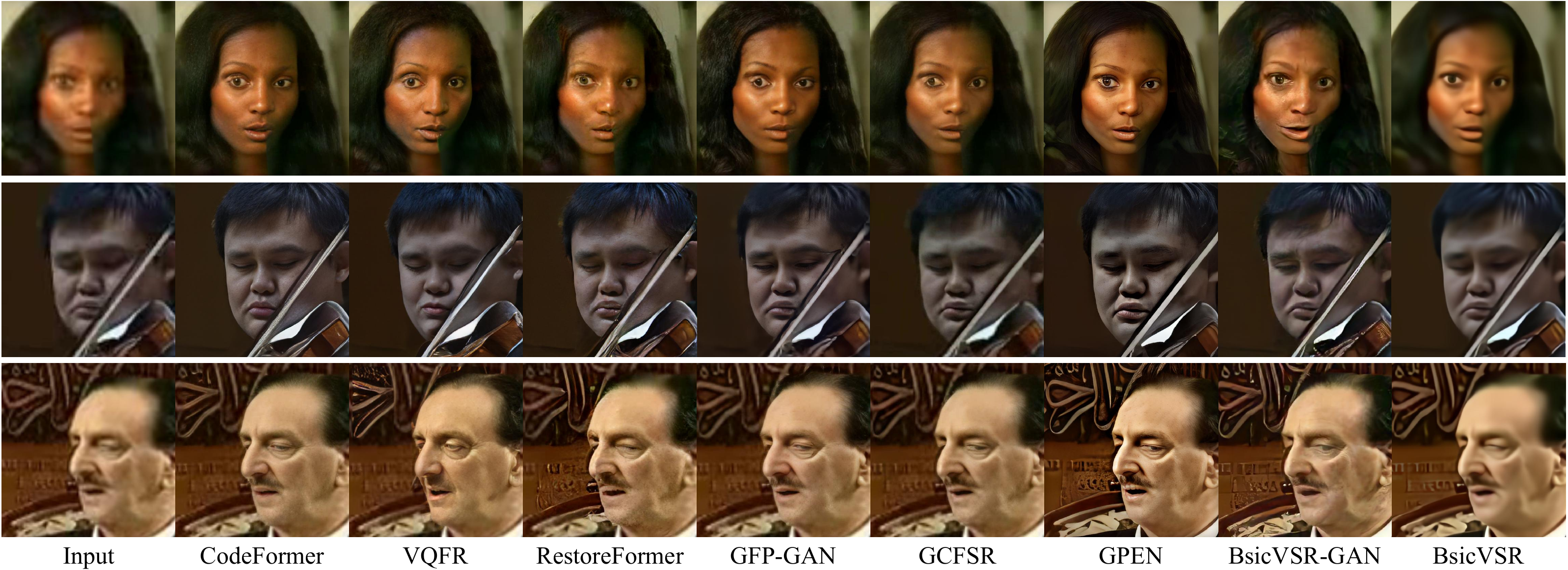}
    \vspace{-2em}
    \caption{Qualitative comparison of both state-of-the-art BFR methods and VSR methods on \textbf{FOS-real}. (\textbf{Zoom in for details})}
  \label{fig:fos_performance_overview}
  \vspace{-1em}
\end{figure*}

\subsection{Subjective Comparison}

\begin{figure}[ht]
    \centering
    \subfloat[Subjective score distributions of 4 BFR methods on \textbf{FOS-real} (\#158) regarding \textit{realness} (blue) and \textit{fidelity} (coral).\label{fig:mos_image}]{%
        \includegraphics[width=1\linewidth]{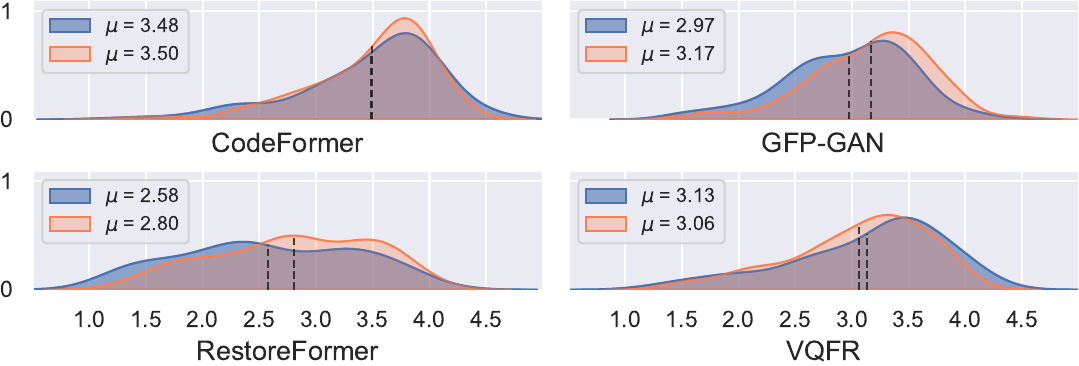}} \\
    \subfloat[Subjective score distributions on 4 BFR methods plus 2 VSR methods on \textbf{FOS-V} (\#108) regarding \textit{reconstruction performance} (orange) and \textit{stability} (green). \label{fig:mos_video}]{\includegraphics[width=1\linewidth]{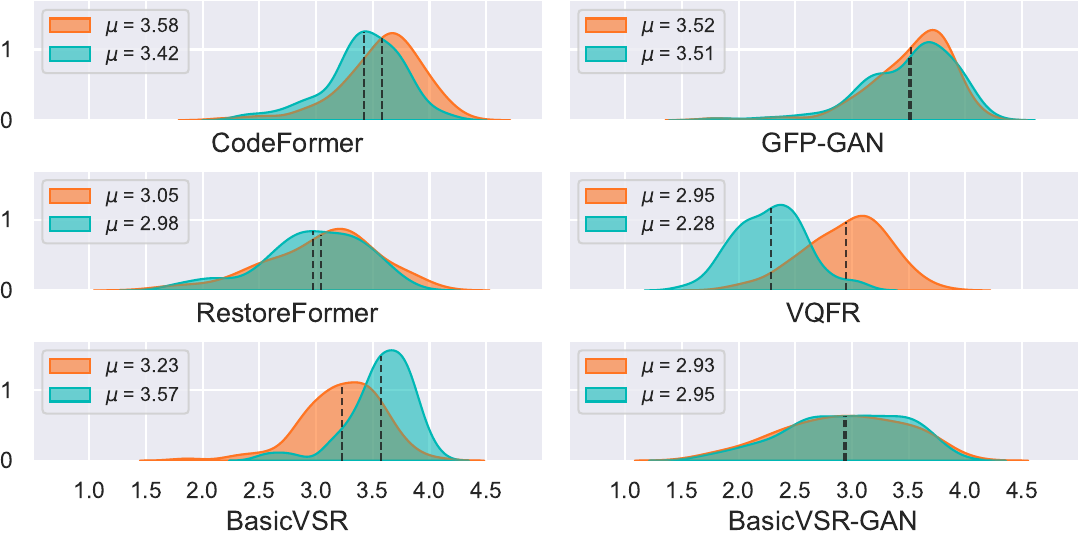}}
    \vspace{-0.5em}
    \caption{The distribution of subjective scores obtained on \textbf{FOS-real} (\#158) and \textbf{FOS-V} (\#108). The mean scores are denoted on the left of each subfigure.}
    \vspace{-2.3em}
\end{figure}

We next conduct a user study to evaluate the most recent 6 BFR and 4 VSR methods.
We select a total of 158 images from \textbf{FOS-real}: 68 images from \textit{full}, 45 images from \textit{occluded}, and 45 images from \textit{side}. 
Our hand-picking ensures diversity by only keeping one face image for one identity. Moreover, we carefully adopt images with diverse facial expressions, poses, and conditions.
For video results comparison, we select a total of 108 video clips from \textbf{FOS-V}. 
The restored results together with the original LQ images, are sent to 28 users for grading in each evaluation dimension. The results of the user study are organized in Figure \ref{fig:mos_image} and Table \ref{tab:user_study_fos_real}. 
The results of GPEN, GCFSR, EDVR, and EDVR-GAN are in the supplementary files due to the space limit.

From Table \ref{tab:user_study_fos_real},
CodeFormer achieves overall great  subjective scores for both \textit{realness} and \textit{fidelity} on \textbf{FOS-real}(\#158): $>3.5$ points on \textit{full} subeset; $>3.0$ points on \textit{occluded} and \textit{side} subsets. These results indicate that CodeFormer yields superior reconstruction performance for \textit{full} subset while showing robustness on other occasions.

\begin{table}[]
    \vspace{-1em}
    \footnotesize
    \centering
    \setlength\tabcolsep{2pt}
    \caption{The subjective scores achieved by 4 BFR methods on \textbf{FOS-real}(\#158) for each subset. Point $\geq$ 3.5 is marked as {\colorbox{myred}{red}}; point $\geq$ 3 is marked as {\colorbox{myblue}{blue}}.}
    \vspace{-0.5em}
    \begin{tabular}{@{}ccccccc@{}}
    \toprule
              & \multicolumn{2}{c}{F.}                                      & \multicolumn{2}{c}{O.}                                      & \multicolumn{2}{c}{S.}                                      \\ \midrule
              & Real.↑                       & Fidel.↑                      & Real.↑                       & Fidel.↑                      & Real.↑                       & Fidel.↑                      \\
    CodeFormer~\cite{codeformer}    & \cellcolor[HTML]{F2DCDB}3.64 & \cellcolor[HTML]{F2DCDB}3.64 & \cellcolor[HTML]{DCE6F1}3.43 & \cellcolor[HTML]{DCE6F1}3.47 & \cellcolor[HTML]{DCE6F1}3.29 & \cellcolor[HTML]{DCE6F1}3.31 \\
    RestoreFormer~\cite{restoreformer} & 2.72                         & 2.90                         & 2.57                         & 2.83                         & 2.36                         & 2.63                         \\
    VQFR~\cite{vqfr}          & \cellcolor[HTML]{DCE6F1}3.42 & \cellcolor[HTML]{DCE6F1}3.32 & 2.92                         & 2.89                         & 2.91                         & 2.85                         \\
    GFP-GAN~\cite{gfpgan}       & \cellcolor[HTML]{DCE6F1}3.05 & \cellcolor[HTML]{DCE6F1}3.22 & 2.83                         & \cellcolor[HTML]{DCE6F1}3.04 & \cellcolor[HTML]{DCE6F1}3.00 & \cellcolor[HTML]{DCE6F1}3.23 \\ \bottomrule
\end{tabular}
    \vspace{-1.5em}
    \label{tab:user_study_fos_real}
\end{table}

While most methods obtain better results in \textit{realness} than \textit{fidelity}, VQFR shows an obvious advantage in \textit{realness} over \textit{fidelity}.
Additionally, VQFR mainly performs well ($>3.0$) on \textit{full} subset but less effective on \textit{occluded} and \textit{side} ($<3.0$).
GFP-GAN exhibits robust performance by obtaining $>3.0$ point in most cases.
As shown in Figure \ref{fig:mos_image}, we find that the subjective score distribution of CodeFormer is more compact while that of RestoreFormer is rather flat, which implies their robust and less-than-robust performance on the test set, respectively.
In general, the average score of all methods is less than 3.5 points in \textit{occluded} and \textit{side} cases, indicating that the robustness of existing BFR methods still needs to be improved.

From Figure \ref{fig:mos_video}, BasicVSR exhibits the highest \textit{stability} performance (point $3.57$) among all methods and shows a small variance in subjective scores. However, its reconstruction only achieves less than $3.3$ points, indicating a gap between the restoration performance and stability maintenance.
CodeFormer achieves the best reconstruction performance (3.58 points) but performs less well than BasicVSR in stability.
Specifically, there exists a huge gap between the reconstruction and stability performance of VQFR.
BFR methods can generate visually pleasant results but underperform in maintaining temporal consistency. This inspires us to combine the advantages of BFR and VSR methods to achieve more stable and realistic video face restoration.

\section{Conclusion}\label{sec:conclusion}
In this work, we propose new benchmark datasets and present a comprehensive benchmark study of the state-of-the-art BFR and VSR methods. The established benchmark dataset \textbf{FOS} consists of face samples mainly from videos, which involve more complex scenarios than existing test datasets. The benchmarking results posed new challenges and identified the future direction of restoration on face videos. Based on our evaluation and analysis, we present the overall conclusions below, which we hope will shed some light on future FVR advances. 
1) The current BFR methods have difficulty generalizing to cases in complex scenarios, such as faces with large pose movements or object occlusion. This generalization issue makes extending BFR methods to VFR solutions more challenging.
2) A more balanced trade-off is expected in future VFR solutions. None existing BFR or VSR methods produce high-quality faces while maintaining iter-frame stability when restoring video faces. 
3) We narrow the gap between the current qualitative evaluation and the subjective scores. The effectiveness of the commonly used evaluation metrics is revisited and studied by leveraging a user study. The results show that NIQE and BRISQUE, are inconsistent with the human eye. Meanwhile, the recently proposed state-of-the-art IQA/FIQA metrics are potential candidates for face quality evaluation.

The collected FOS datasets may have negative social impacts such as privacy leaking. To mitigate the influence of privacy, the data are derived from two public datasets and self-collected internet data which involves mainly celebrities. Any user who acquires the datasets must follow the license provided by \cite{vggface2}.

\vspace{-0.5em}
\section*{Acknowledgements}
\vspace{-0.3em}
This work was supported by the National Natural Science Foundation of China (Grant No.62276251), and the Joint Lab of CAS-HK.
{
\vspace{-2em}
    \small
    \bibliographystyle{ieeenat_fullname}
    \bibliography{main}
}


\end{document}